%% file: main.tex
\documentclass[letterpaper, 10 pt, conference]{ieeeconf}  

\usepackage{xcolor}
\usepackage[nolist]{acronym} 
\usepackage{graphicx}
\usepackage{cite}

\usepackage{subcaption}
\usepackage{amsmath}
\usepackage{amsfonts}

\IEEEoverridecommandlockouts                              

\usepackage{url}
\input{acronyms.tex}

\title{\LARGE \bf
Enabling Robot Manipulation of Soft and Rigid Objects with Vision-based Tactile Sensors 
}

\author{Michael C. Welle*$^{1}$, Martina Lippi*$^{2}$, Haofei Lu$^{1}$, Jens Lundell$^{1}$, Andrea Gasparri$^{2}$, Danica Kragic$^{1}$
\thanks{*Contributed equally.}
\thanks{ ${}^1$KTH Royal Institute of Technology Stockholm, Sweden, {\it\small \{mwelle,haofeil,jelundel,dani\}@kth.se}}
\thanks{ ${}^2$Roma Tre University, Rome, Italy  {\it\small \{martina.lippi,andrea.gasparri\}@uniroma3.it} }
\thanks{This work was supported by the Swedish Research Council, Knut and Alice Wallenberg Foundation, by the European Research Council (ERC-884807), and by the European Commission (Project CANOPIES-101016906).}
}

\begin{document}

\maketitle
\thispagestyle{empty}
\pagestyle{empty}

\begin{abstract}
Endowing robots with tactile capabilities opens up new possibilities for their interaction with the environment, including the ability to handle fragile and/or soft objects.
In this work, we equip the robot gripper with low-cost vision-based tactile sensors and propose a manipulation algorithm that adapts to both rigid and soft objects without requiring any knowledge of their properties. The algorithm relies on a touch and slip detection method, which considers the variation in the tactile images with respect to reference ones. 
We validate the approach on seven different objects, with different properties in terms of rigidity and fragility, to perform unplugging and lifting tasks. Furthermore, to enhance applicability, we combine the manipulation algorithm with a grasp sampler for the task of finding and picking a grape from a bunch without damaging~it.  
\end{abstract}

\input{includes/intro.tex}

\input{includes/rw.tex}

\input{includes/method.tex}

\input{includes/experiments.tex}

\input{includes/discussion_conclusion.tex}

\bibliographystyle{IEEEtran}
\bibliography{refs}

\end{document}

%% file: acronyms.tex
\newacro{dnn}[DNN]{Deep Neural Network}
\newacro{cnn}[CNN]{Convolutional Neural Network}
\newacro{dl}[DL]{Deep Learning}
\newacro{ml}[ML]{Machine Learning}
\newacro{ekf}[EKF]{Extended Kalman Filter}

%% file: includes/intro.tex
\section{Introduction}

Humans have a remarkable ability to manipulate a wide range of objects in various situations. For instance, we can effortlessly pick up and hold objects of different shapes and sizes, adjust our grip, and move them around without much thought. Furthermore, we can manipulate \emph{fragile} objects without damaging them.  This level of flexibility and adaptability cannot be easily replicated in robotic systems~\cite{billard2019trends}.
A major challenge for current robotic systems is the deficiency in sensory capabilities when compared to humans: 
human fingers are soft, sensitive tactile sensors that enable humans to feel contact forces as well as detect slippage of objects with ease.
Equipping robotic hands or grippers with similar tactile sensing capabilities has been a long-standing goal of the research community \cite{survey_tactile}. 

Recently,
vision-based tactile sensors have received increased popularity in the community \cite{vision_tactile_survey} thanks to their relatively low cost, high resolution, and ease of use. Examples of vision-based tactile sensors are GelSight \cite{yuan2017gelsight}, TacTip \cite{ward2018tactip}, and DIGIT \cite{lambeta2020digit}.
The basic principle of these sensors is to use a camera to capture the deformation of an elastomer material occurring in case of contact with objects.
By combining this vision-based tactile sensing with powerful machine learning techniques, researchers have managed to demonstrate impressive tasks such as in-hand manipulation \cite{lambeta2020digit} and ball rolling \cite{tian2019manipulation}  (see Sec. II for more details).
However, these solutions typically rely on large neural networks that may be tailored to specific sensors and objects and may require re-training to address different tasks. Additionally, many of the existing approaches are designed to handle rigid objects. 

\begin{figure}
    \centering
    \includegraphics[width=\linewidth]{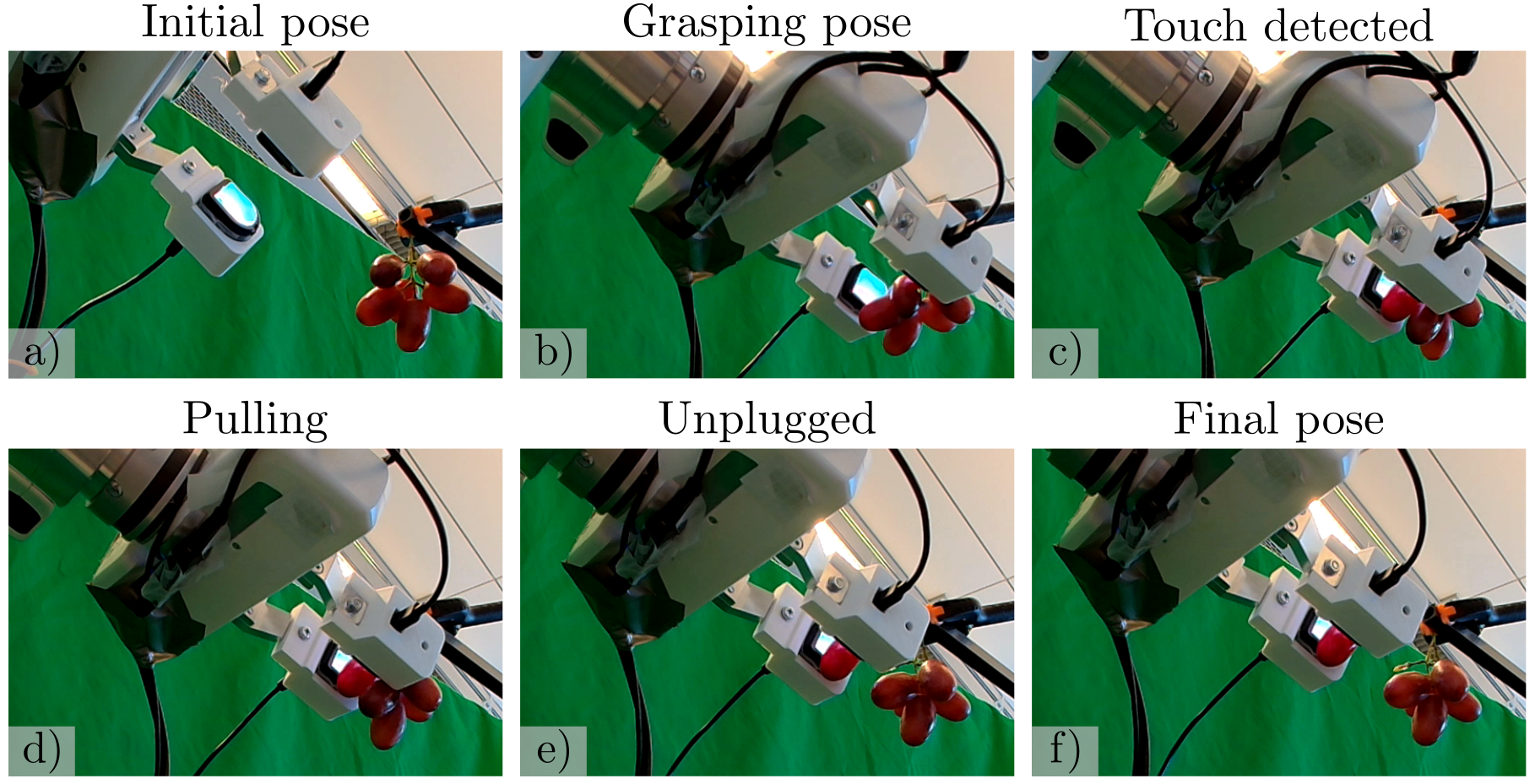}
    \caption{Sequence of states for picking a grape berry from a bunch, without damaging it, using visual-tactile sensors mounted on the robot gripper. 
    }
    \label{fig:real_bunch}
\end{figure}

In this work, we aim to bridge this gap by presenting a simple learning-free touch and slip detection technique for vision-based tactile sensors that relies  on a single hyperparameter, which can be easily tuned without gathering large datasets.
We design an algorithm that combines such detection methods to handle a wide range of objects,  from soft to rigid ones, without requiring any  prior knowledge of the object properties. 
Finally, we integrate this algorithm into a grasping pipeline \cite{zehang_pp} that proposes a set of grasp poses from which we select one to execute for picking a grape from a bunch, as shown in Fig. \ref{fig:real_bunch}. 
In detail, our contributions are:
\begin{itemize}
    \item A simple-to-deploy touch and slip detection method; 
    \item The design of an algorithm to manipulate soft or rigid objects, which can potentially be fragile;  
    \item Extensive real-world experimental validation using low-cost DIGIT sensors \cite{lambeta2020digit} with seven different objects, such as tomatoes and ridged connectors (all videos on the project website\footnote{\label{fn:website} \url{https://vision-tactile-manip.github.io/exp/}}); 
    \item  Deployment of the framework in a comprehensive grasping pipeline detaching table-grapes from a bunch with an off-the-shelf robotic platform. 
\end{itemize}

%% file: includes/rw.tex
\section{Related Work}

Here, we review prior work in vision-based tactile manipulation. For an overview of non-vision-based tactile sensing and manipulation, we refer the reader to the following two reviews \cite{li2020review,kappassov2015tactile}.

The general interest in vision-based tactile manipulation took off with the invention of the GelSight sensing technology \cite{yuan2017gelsight}, and, in particular, its adaption to fit gripper form factors \cite{wang2021gelsight, donlon2018gelslim,taylor2022gelslim,ma2019dense}. Since conception, these and other similar types of sensors have been used in a range of tasks, including pose estimation \cite{villalonga2021tactile,li2014localization}, grasping \cite{CalandraOUYLAL17,calandra2018more,hogan2018tactile,yamaguchi2017implementing}, slip detection \cite{dong2017improved}, shape reconstruction \cite{bauza2019tactile,suresh2022shapemap}, object tracking \cite{izatt2017tracking}, and closed-loop control \cite{tian2019manipulation,she2021cable,wang2020swingbot,dong2019tactile}. 

Most of the above works \cite{CalandraOUYLAL17, calandra2018more, hogan2018tactile,villalonga2021tactile,suresh2022shapemap,bauza2019tactile,tian2019manipulation,she2021cable,wang2020swingbot,dong2019tactile}, tackles the problem with data-driven methods. For instance, in grasping, Calandra et al. \cite{CalandraOUYLAL17} used a \ac{cnn} to classify grasp success given an RGB and tactile image of the grasped object, while the authors of \cite{calandra2018more, hogan2018tactile} used a \ac{cnn} to suggest local gripper adjustments based on the current tactile image for improving grasps. Other example works in vision-based tactile deformable object manipulation have explored closed-loop control of the gripper width for cable manipulation \cite{she2021cable} and pivoting \cite{wang2020swingbot}. From an application perspective, the main limitation of the data-driven methods is gathering datasets diverse enough for training models that can generalize to many unseen objects, which is both time-consuming and non-trivial to realize. 

To mitigate the problem inherent to data-drive vision-based tactile manipulation methods, a few works have explored learning-free approaches \cite{li2014localization,dong2017improved,izatt2017tracking}. The earliest of these works, \cite{li2014localization}, did tactile pose estimation of a USB stick and used the estimated pose for inserting the USB into a port. Similar to \cite{li2014localization}, Izatt et al. \cite{izatt2017tracking} proposed an \ac{ekf}-based method for tracking the pose of a manipulated object using point clouds from an RGB-D camera and the tactile image from a GelSight sensor. 

Out of the learning-free methods, the one most similar to ours is \cite{dong2017improved}, where the authors also proposed an algorithm for adjusting the grasp force to avoid object slippage when grasping. That work, however, required markers on the GelSight sensor to detect slippage and  mostly evaluated the method on rigid objects. Our method, on the other hand, is marker-free and can handle rigid, deformable, and \textit{fragile} objects. Moreover, it is the first vision-based tactile manipulation method to successfully realize cable unplugging.

%% file: includes/method.tex
\section{Method}
As mentioned in the introduction, we consider a robot equipped with two vision-based tactile sensors mounted on its gripper, as shown in Fig. \ref{fig:real_bunch}. Our focus is on designing a manipulation strategy that adaptively modulates the gripper closure in order 
to manipulate an object of interest, which might either be soft or rigid and potentially fragile.
Specifically, the objective is to ensure that the gripper can securely grasp the object without losing it during the manipulation task and,  at the same time, it does not compress the object more than necessary to hold it. The latter aspect is fundamental in the case of fragile objects, such as table-grapes, where over-compression may result in damage to the objects. 
Examples of manipulation tasks that require such ability include unplugging or detaching objects, which require a certain amount of force to be exerted without over-compressing the object, and moving fragile items, where it is crucial to modulate the gripper closure carefully, as mentioned above.
As tactile sensors, we use DIGIT sensors, but the approach can be generalized to any vision-based tactile sensor.

In order to successfully modulate the gripper closure and perform a manipulation task on both soft and rigid objects, 
we identify two main capabilities that the system must provide: 
\begin{enumerate}
    \item Touch detection, which establishes whether contact with an external object has occurred, regardless of the texture, shape, rigidity, and other physical properties of the object; 
    \item Slip detection, which establishes, after a contact detection,  whether the object is sliding out of the gripper. 
\end{enumerate}
Based on these capabilities, a manipulation algorithm is designed which establishes the robot reaction in terms of motion and modulation of the gripper closure in response to either touch or slip detection.
Desirable features to achieve such capabilities are sensitiveness and ease of tuning. Specifically, for touch detection, we want the system to be sensitive enough to detect contact, even with soft objects, without  over-compressing them. Nevertheless, a balance must exist that allows the system to be robust with respect to noisy data of the tactile sensor, and  avoid misclassifying such noise as a contact state.
Regarding slip detection, it should be sensitive enough to prevent the object from completely slipping out of the gripper, but at the same time, it should not be overly sensitive and detect slip for minor changes in the object's positioning.
Our approach addresses both capabilities with the same algorithm and only relies on a single hyperparameter that is easy to tune.
In the following, we first describe the touch and slip detection method and then describe how to exploit it into a manipulation algorithm.

\subsection{Touch and Slip Detection}

\begin{figure}
    \centering
    \includegraphics[width=\linewidth]{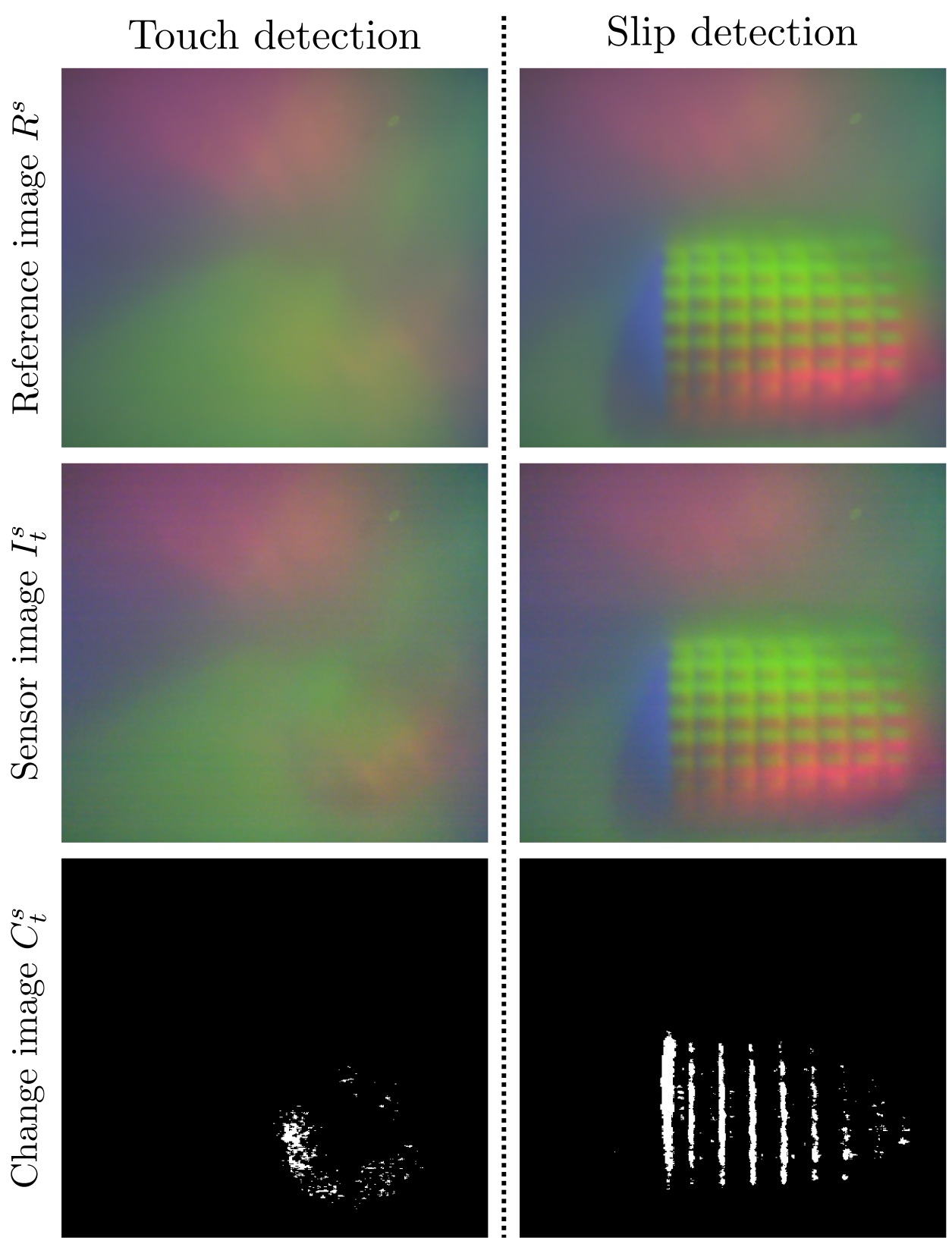}
    \caption{Examples of touch detection on a grape (left column) and slip detection on an AUX connector (right column)}
    \label{fig:touch_and_slip}
\end{figure}

The basic principle behind the touch and slip detection methods is the same. Given a reference image of a tactile sensor, which is updated over time according to the manipulation algorithm,  we evaluate the pixel-wise variation of consecutive frames with respect to the reference image. 
If this variation is above a certain threshold, a \emph{change} in the tactile sensor state is recorded. Next, the change detection is  \emph{categorized} as either touch or slip detection. 
In the following, firstly, we  focus on how to recognize a change in the sensor state and, secondly, we describe how the classification into touch or slip is made. Note that these detections  are carried out  independently for each sensor and then combined in the manipulation algorithm. 

Let $I^s_t\in\mathbb{R}^{h\times w \times 3}$ be the normalized tactile input image (i.e., values in the range $[0,1]$), with height $h$ and width $w$, of the sensor $s$ at time step $t$, where  \mbox{$s\in\{1,2\}$}, and let $R^s \in \mathbb{R}^{h\times w \times 3}$ be the normalized reference image related to the same sensor. 
 Then, we calculate the RGB difference image $D^s_t\in\mathbb{R}^{h\times w \times 3}$ as
\begin{equation}
   D^s_t = |I^s_t - R^s|,    
\end{equation}
 which is converted into a mono-dimensional image \mbox{$\bar{D}^s_t\in \mathbb{R}^{h\times w}$} by averaging the three (RGB) channels. At this point,   a thresholding operation is applied to mitigate the noise in the tactile sensor data. Specifically,
we assign  zero value to pixels with intensity below a noise threshold $\tau_n^s\in[0,1]$, and a value of $1$  otherwise. The result is a binary difference image  \mbox{$\widetilde{D}_{t}^s\in\mathbb{B}^{h\times w}$}, with $\mathbb{B}=\{0,1\}$, 
capturing the relevant difference of the current tactile image compared to the reference image. 
Such a binary image is combined with the one at the previous time step to detect a change in the sensor state. Specifically, we define the change image $C_t^s\in\mathbb{B}^{h\times w}$  at time $t$ as 
\begin{equation}
    C_t^s = \widetilde{D}^{s}_{t-1} \odot \widetilde{D}_{t}^s,
\end{equation}
where $\odot$ denotes the element-wise multiplication (Hadamard product). Then, we rely on the fulfillment of the following condition to determine whether a change has occurred 
\begin{equation}
   \frac{1}{wh}\sum_{i=1}^h\sum_{j=1}^w C_t^s(i,j) \geq \tau_d,
\end{equation}
with $\tau_d\in[0,1]$ a detection threshold, i.e., a change occurs 
if the ratio of ones in $ C_t^s$  to the total number of pixels exceeds a specified threshold. Note that the use of consecutive images in $ C_t^s$ enhances the system robustness since it enables disregarding pixels that are equal to one for a single time step.  blue Such occurrences can arise due to the presence of noise in the vision-based tactile sensor.

 After detecting a change, we classify it by simply analyzing when it has occurred. Specifically, since at the start, we assume no contact  with the object, when the first change is recognized, this is classified as touch detection. On the other hand, if the change occurs  after a touch has been established,  this is classified as slip detection. Note that in the first case, the reference image is related to the ``empty" sensor, i.e., no object in contact, while in the second case, the reference image represents the image obtained when holding the object. More details on the reference image update are provided in the next section, along with the manipulation algorithm. Fig.~\ref{fig:touch_and_slip} (left column) shows an example of images obtained in case of  touch detection with a grape, where  Fig.~\ref{fig:touch_and_slip} (top left) represents the reference (``empty") image,  Fig.~\ref{fig:touch_and_slip} (middle left) shows the current tactile image and Fig.~\ref{fig:touch_and_slip} (bottom left) reports the change image. 
Similarly, Fig.~\ref{fig:touch_and_slip} (right column) shows examples of images obtained in case of slip detection of an AUX connector. 

Note that  two thresholds are used in the above for detection, i.e., the noise threshold $\tau_n^s$ and the detection threshold $\tau_d$.  
To determine the first threshold $\tau_n^s$, we propose an easy  automatic calibration procedure for each sensor $s$ to be run when there is no contact. In detail, we  record the maximum values in the averaged difference image for $K$ time steps and then compute the mean value, i.e, 
\begin{equation}
\tau_n^s=\frac{1}{K}\sum_{t=1}^{K} \max(\bar{D}^s_t). 
\end{equation}
It follows that the only hyperparameter to select for the touch/slip detection is $\tau_d$ as detailed in the following.

\subsection{Manipulation Algorithm} 

\begin{figure}
    \centering
    \includegraphics[width=1\linewidth]{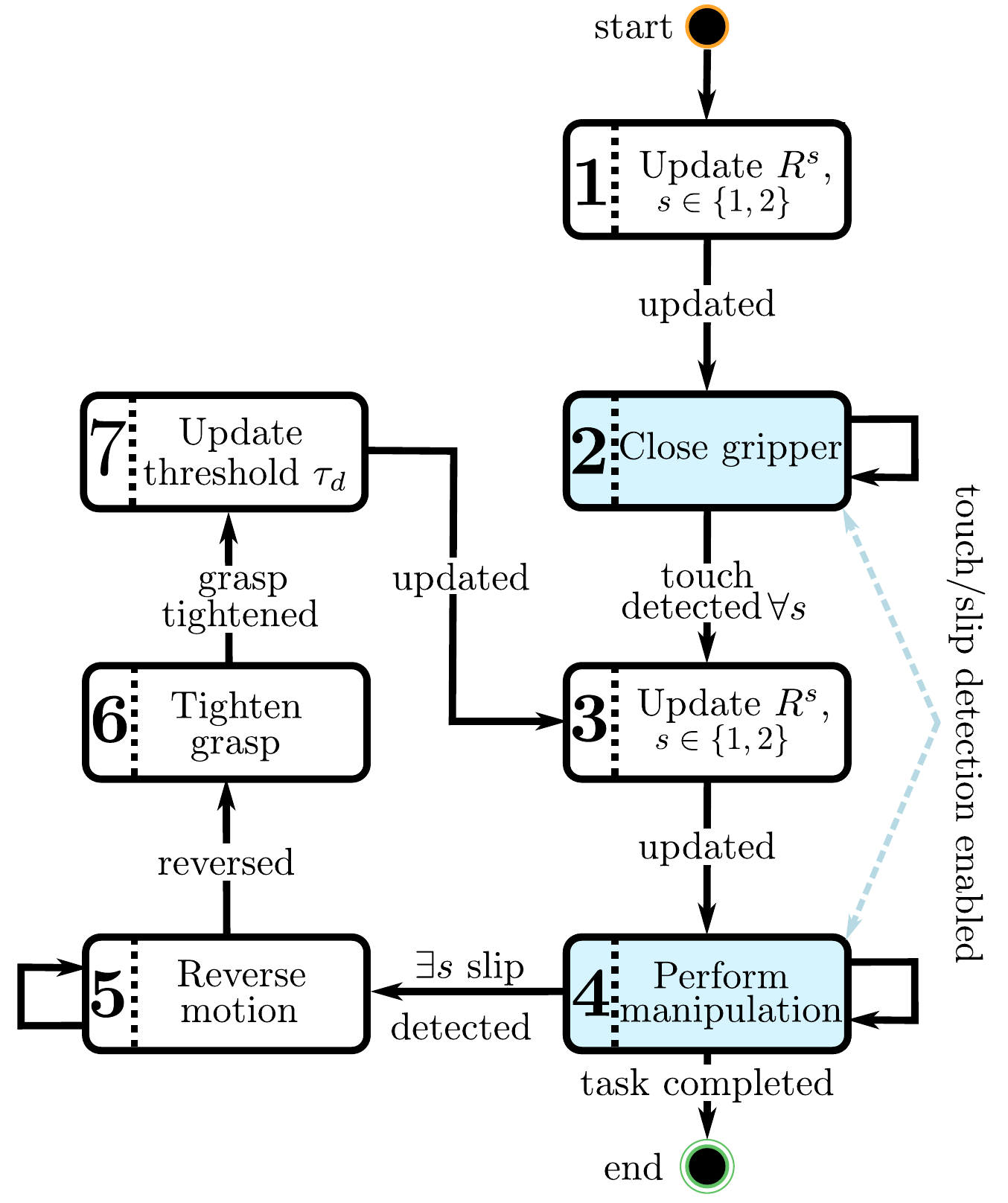}
    \caption{Overview of the manipulation algorithm. 
    The states where the touch/slip detection is active are highlighted in blue color. 
    }
    \label{fig:algo}
\end{figure}

 Fig. \ref{fig:algo} presents an overview of the proposed manipulation algorithm. Briefly, the basic idea is that once touch detection is complete, the robot can initiate the manipulation motion. If any slip occurs, the gripper closure is adjusted to ensure a firm grasp and the manipulation motion can be continued.

At the start, we assume that the robot is in a pre-grasping pose, wherein the closure of the gripper will result in the grasping of the object. Additionally, we assume that the noise thresholds $\tau_n^s$ have been calculated according to the calibration procedure described earlier.
No knowledge is assumed regarding the object to manipulate in terms of  texture, shape, rigidity, and other physical properties.

The first objective is to retrieve the initial reference images  $R^s$ of both the tactile sensors, i.e., $s\in\{1,2\}$ (state 1 in the Fig. \ref{fig:algo}). Such reference images are representative of the tactile images when no contact is happening, as shown for instance in Fig.~\ref{fig:touch_and_slip} (top left).  
These are obtained by collecting and averaging $N$ tactile images for each sensor, i.e.,
\begin{equation}\label{eq:ref}
    R^s= \frac{1}{N}\sum_{t=1}^{N} I^s_t, \,\quad s\in\{1,2\}. 
\end{equation}
Once the reference images are defined, touch detection is  activated on both sensors (indicated with blue color), and the gripper begins to close gradually (state 2 in the figure).  The gripper keeps closing until both sensors detect the occurrence of touch, in order to ensure that the object can be held by the two fingers of the gripper.

At this point, the reference images $R^s, \forall s,$  are updated again to enable slip detection. As in \eqref{eq:ref}, we gather the next $N$ tactile  images and average them for each sensor (state 3 in Fig.~\ref{fig:algo}). In this case, the reference images are representative of the tactile images when the robot is holding the object, as shown for instance in Fig.~\ref{fig:touch_and_slip} (top right).  Therefore, a variation with respect to the reference image would indicate a slip of the object. 

After completing the update of the reference images, the robot starts the manipulation and moves as required (state 4 of Fig.~\ref{fig:algo}). For instance, in an unplugging task, the robot would begin moving in the unplugging direction. While the manipulation is in progress, slip detection is activated.

If slip is detected by \emph{either} of the tactile sensors, we apply a reaction strategy. Our first step is to reverse the manipulation motion, meaning that the robot  returns to the configuration before slip detection (state 5 of the figure).  In contact-rich manipulation tasks, such as unplugging objects, this action helps to prevent the object from slipping out of the gripper completely in the event that slip conditions persist over multiple time steps. However, in non-contact-rich tasks, such as object transportation, this phase might be unnecessary.  

Once the reverse motion is complete,  the gripper closure is modulated and the grasp is tightened (state 6 of Fig.~\ref{fig:algo}).  To this aim, the gripper is closed by a small amount depending on the gripper resolution.  In our case, we utilize displacements of $0.001$~m. The last step of the slip reaction strategy is updating the detection threshold $\tau_d$ (state 7 of the figure). Specifically, we express the threshold as 
$$\tau_d = k_s \tau_{d,ini}$$
where $\tau_{d,ini}$ is the initial value of the threshold that the user has to select, while $k_s\in\mathbb{R}^+$ is a positive gain which is doubled every time a slip is detected, i.e., $k_s = 2\, k_{s,prev} $ where $k_{s, prev}$ is the previous value of $k_s$ and is initialized to~$1$. 
This exponential increase of the threshold allows us to adapt the slip detection sensitiveness over time: the tighter the object is grasped, the higher the tolerance for slip detection must be; otherwise, the system would keep detecting slippage for any minor change, preventing the successful manipulation of the object.  

At the end of the slip reaction strategy, the reference images are updated (state 3 of the figure) and the manipulation is recovered (state 4 of the figure) until task completion.

\subsection{Integration into Grasping Pipeline}\label{sec:integration}

Our manipulation algorithm assumes that a pre-grasping pose is given. In simple scenarios, such as picking objects from a table, using predefined grasp poses is reasonable. However, manually defining successful grasp poses in more challenging scenarios, such as picking grapes from a bunch, is non-trivial. Therefore, in more challenging scenarios, we use the grasp sampling method in \cite{zehang_pp} to suggest good grasp poses for the manipulation algorithm.

In short, the grasp sampler from \cite{zehang_pp} is a conditional generative neural network that takes as conditional inputs an incomplete point cloud of the object to grasp and a reference grasp approach direction. From these inputs, the sampler then generates multiple grasps whose approach direction is close but not exactly the reference one. To simplify the task, 
we select a region of interest given the point cloud and feed it to the grasp sampler. 
In addition, we also manually select, from all generated grasps, the grasp the robot should reach, which is provided to the low-level Cartesian controller. Once the grasp pose is reached, the manipulation algorithm described above is executed.

%% file: includes/experiments.tex
\section{Experiments}

\begin{figure*}[ht]
\def\imgsize{0.137}
\def\imgsizetac{0.137}
\centering
\begin{subfigure}[b]{\imgsize\textwidth}
  \centering
  \includegraphics[width=1\linewidth]
  {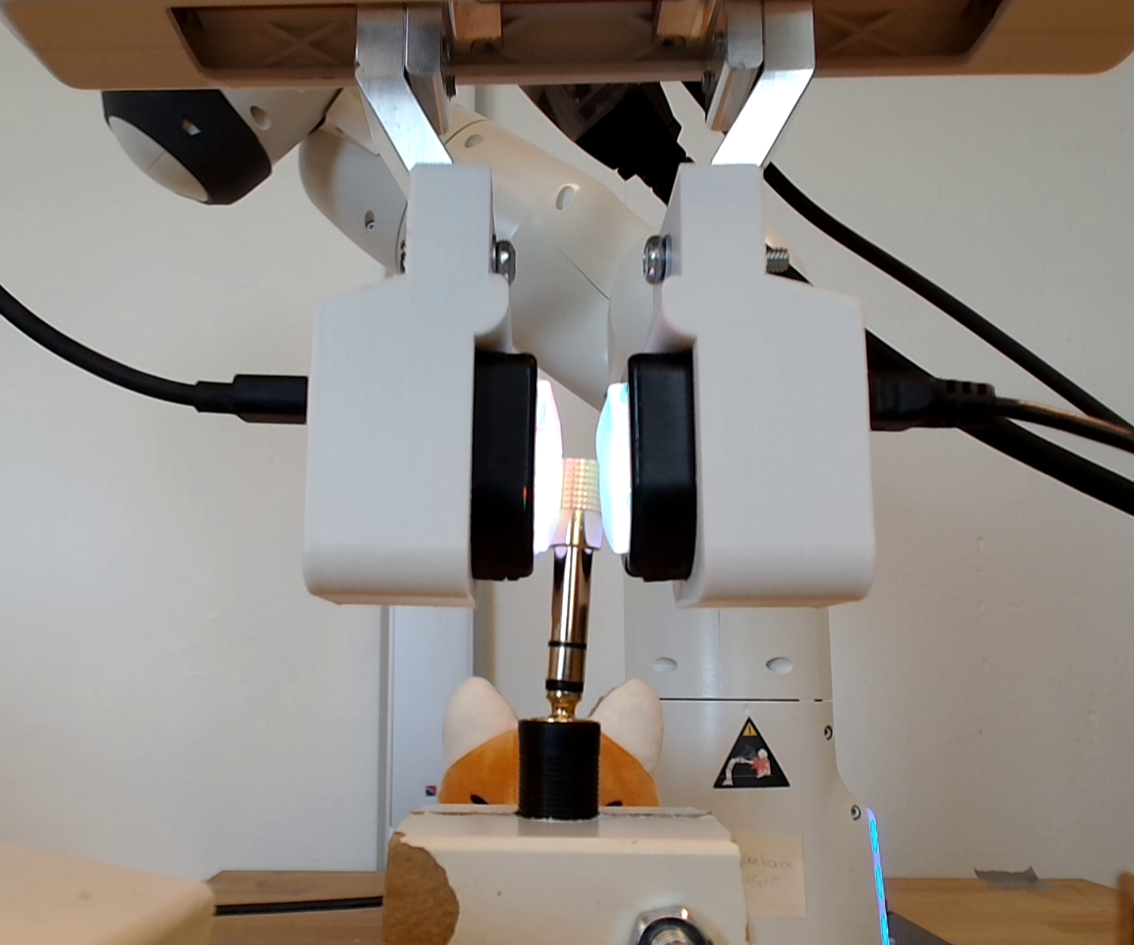}
\end{subfigure}
\begin{subfigure}[b]{\imgsize\textwidth}
  \centering
  \includegraphics[width=1\linewidth]
  {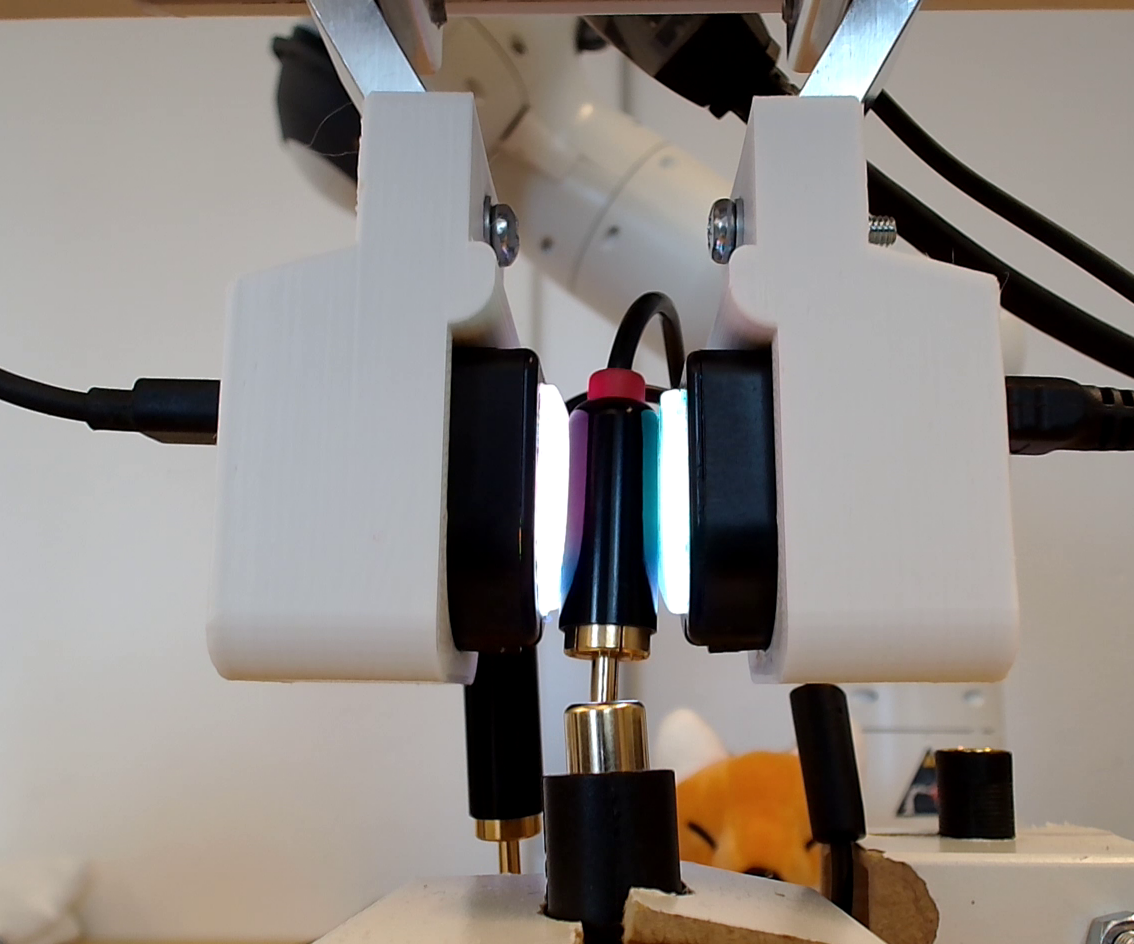}
\end{subfigure}
\begin{subfigure}[b]{\imgsize\textwidth}
  \centering
  \includegraphics[width=1\linewidth]
  {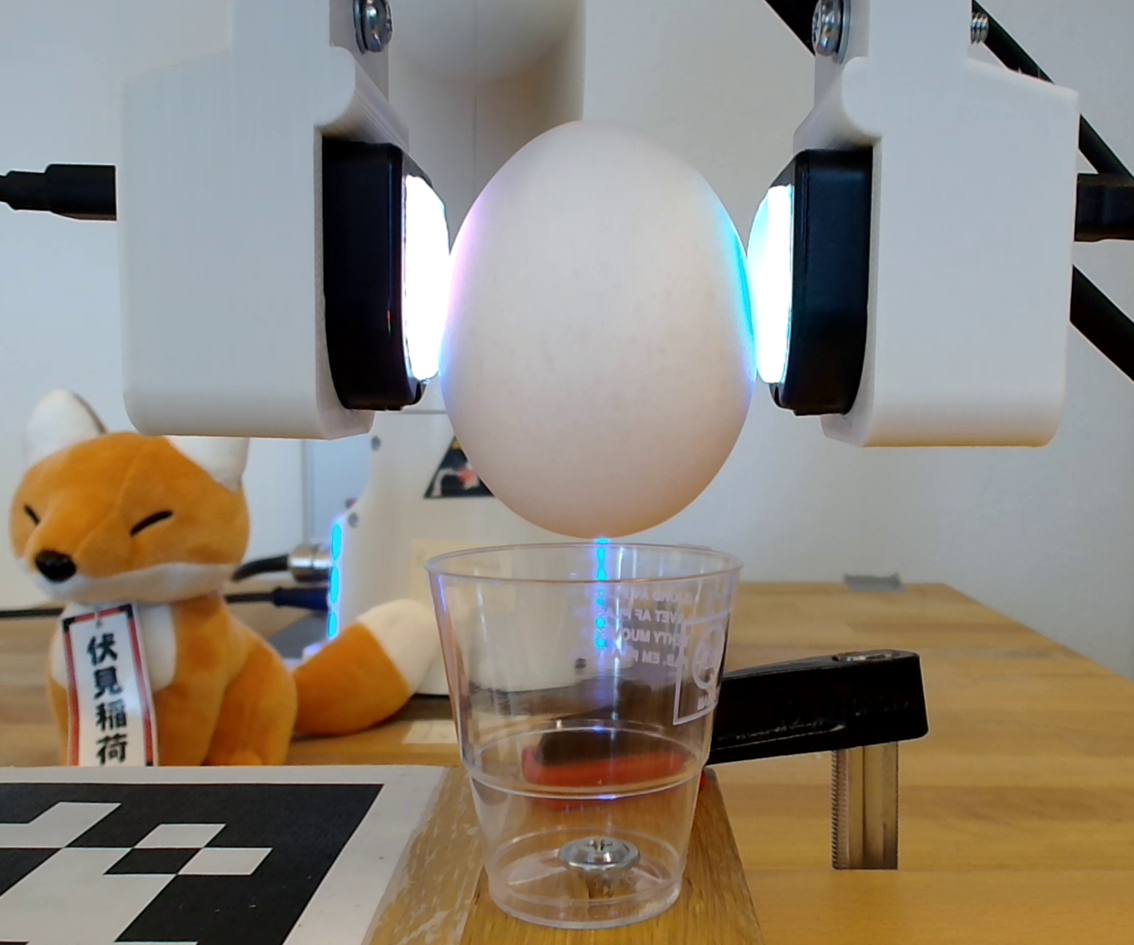}
\end{subfigure}
\begin{subfigure}[b]{\imgsize\textwidth}
  \centering
  \includegraphics[width=1\linewidth]
  {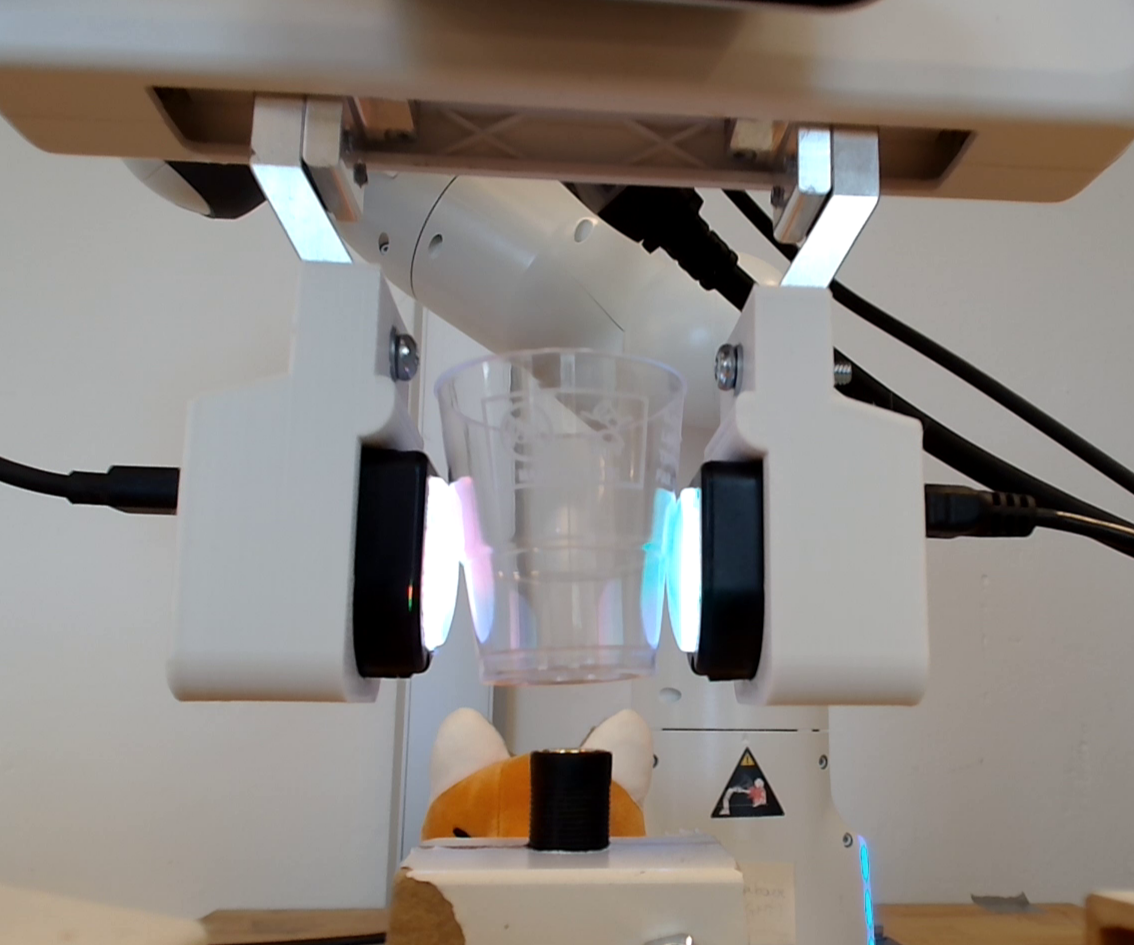}
\end{subfigure}
\begin{subfigure}[b]{\imgsize\textwidth}
  \centering
  \includegraphics[width=1\linewidth]
  {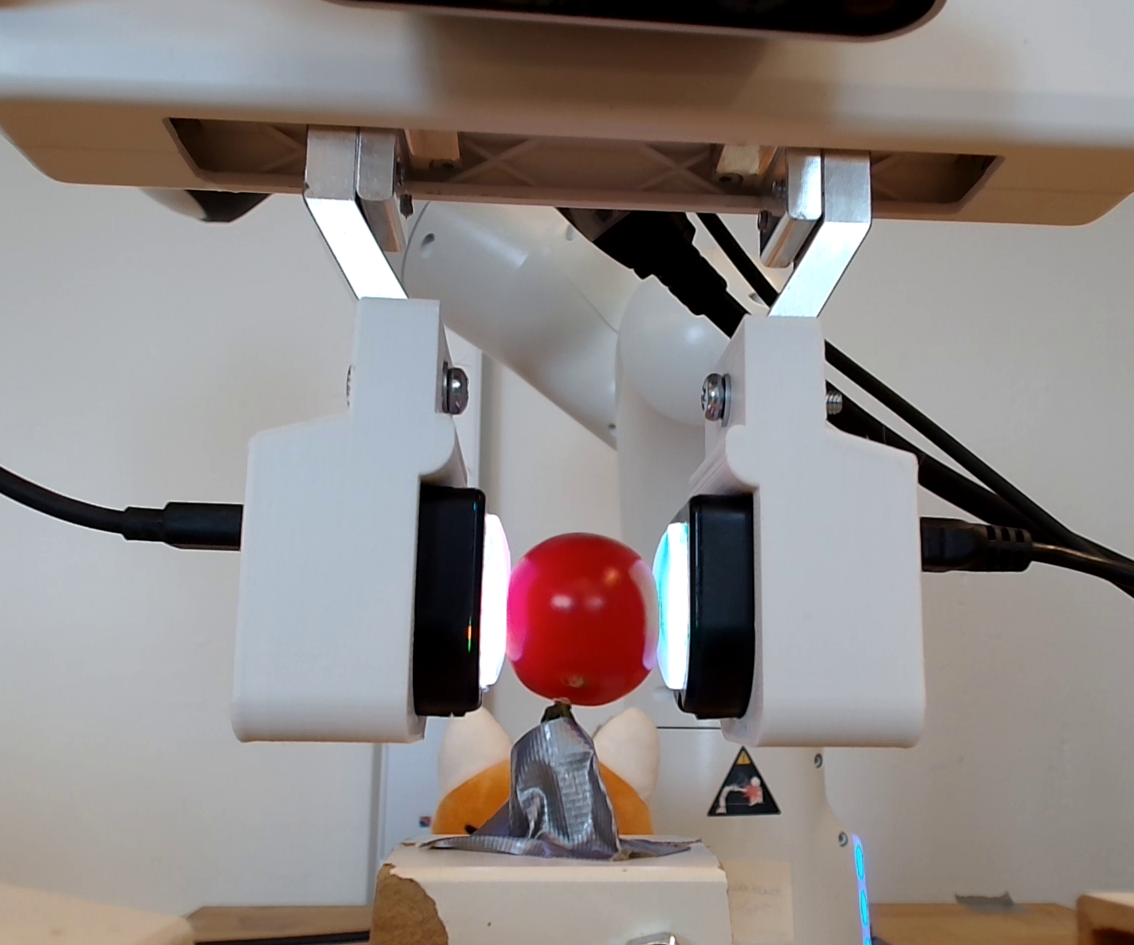}
\end{subfigure}
\begin{subfigure}[b]{\imgsize\textwidth}
  \centering
  \includegraphics[width=1\linewidth]
  {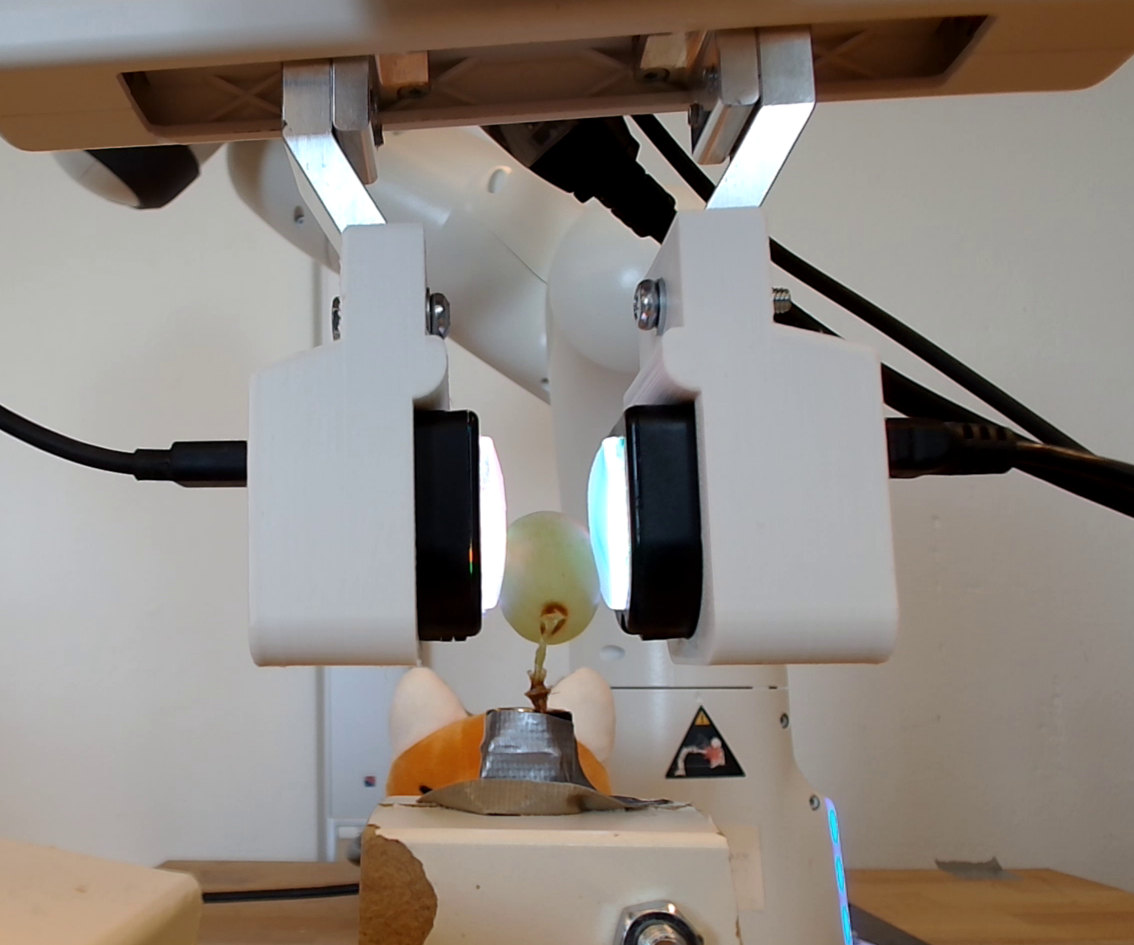}
\end{subfigure}
\begin{subfigure}[b]{\imgsize\textwidth}
  \centering
  \includegraphics[width=1\linewidth]
  {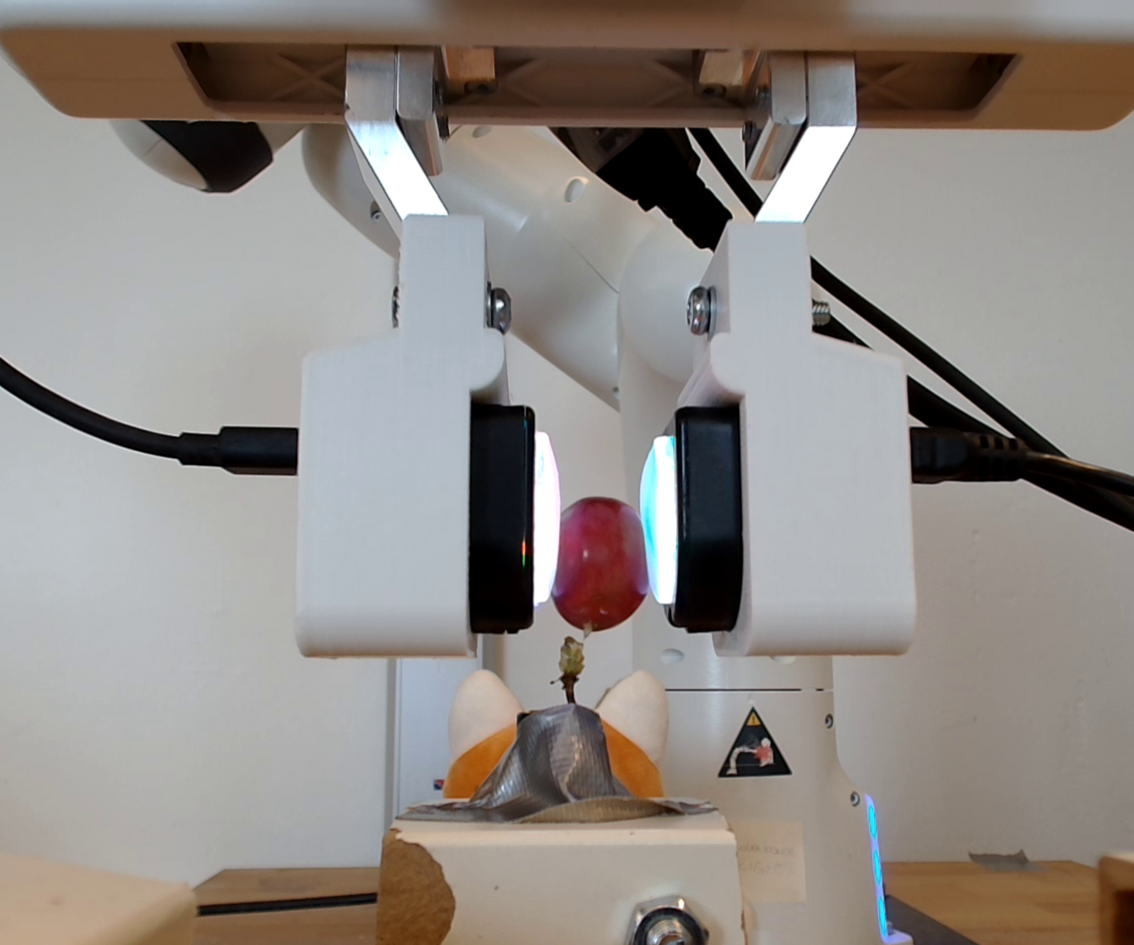}
\end{subfigure}
\\ \vspace{3pt}
%
%
\begin{subfigure}[b]{\imgsizetac\textwidth}
  \centering
  \includegraphics[width=1\linewidth]
  {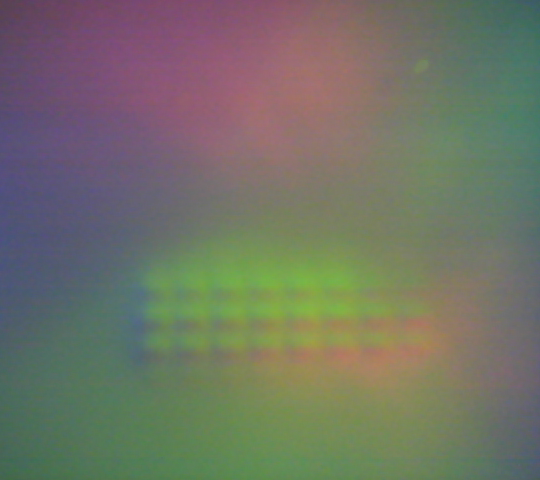}
    \caption{Rough conn. }
    \label{fig:rough_connector}
\end{subfigure}
\begin{subfigure}[b]{\imgsizetac\textwidth}
  \centering
  \includegraphics[width=1\linewidth]
  {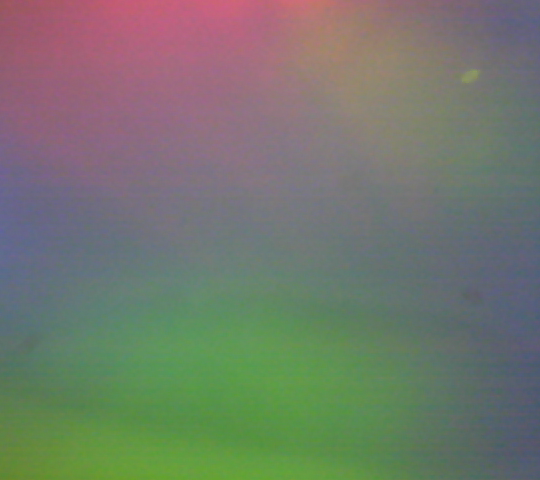}
    \caption{Smooth conn. }
    \label{fig:smooth_connector}
\end{subfigure}
\begin{subfigure}[b]{\imgsizetac\textwidth}
  \centering
  \includegraphics[width=1\linewidth]
  {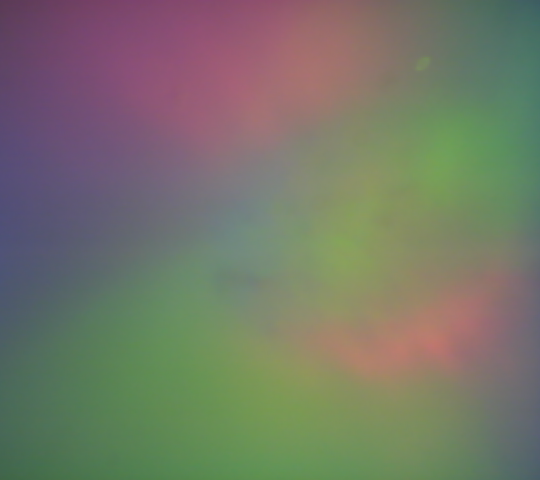}
    \caption{Raw egg. }
    \label{fig:egg}
\end{subfigure}
\begin{subfigure}[b]{\imgsizetac\textwidth}
  \centering
  \includegraphics[width=1\linewidth]
  {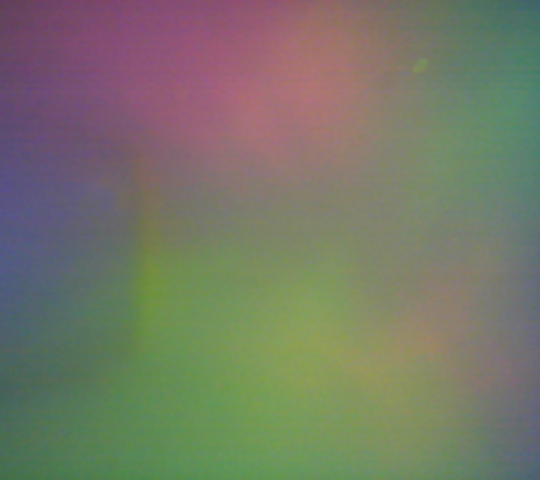}
    \caption{Glass. }
    \label{fig:glass}
\end{subfigure}
\begin{subfigure}[b]{\imgsizetac\textwidth}
  \centering
  \includegraphics[width=1\linewidth]
  {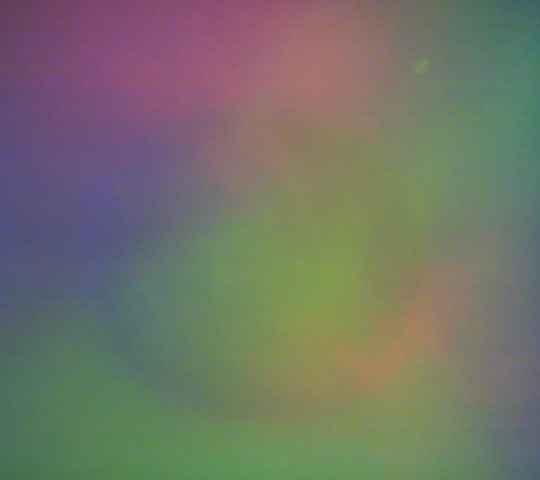}
    \caption{Tomato. }
    \label{fig:tomato}
\end{subfigure}
\begin{subfigure}[b]{\imgsizetac\textwidth}
  \centering
  \includegraphics[width=1\linewidth]
  {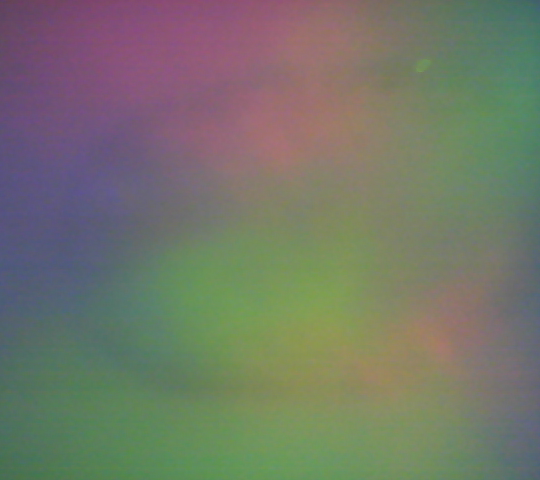}
    \caption{White grape. }
    \label{fig:whitegrape}
\end{subfigure}
\begin{subfigure}[b]{\imgsizetac\textwidth}
  \centering
  \includegraphics[width=1\linewidth]
  {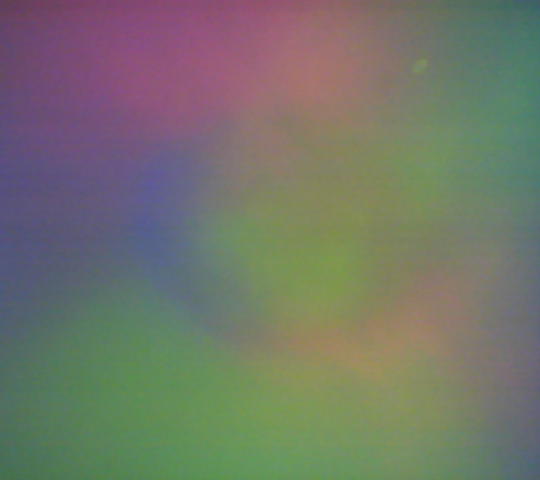}
    \caption{Black grape. }
    \label{fig:blackgrape}
\end{subfigure}
\caption{Illustrations of the considered objects to manipulate (top row) and the respective tactile sensor images (bottom row). }
\label{fig:objects}
\end{figure*}
We validated the effectiveness of the proposed approach with a Franka Emika Panda robot, equipped with two DIGIT sensors on the gripper, performing various manipulation tasks, as shown in Fig. \ref{fig:objects}. 
In detail, we performed an extensive validation of the manipulation algorithm on $7$ different objects, including rigid, soft, and fragile ones. Finally, inspired by the needs of the European Project CANOPIES, focusing on the precision agriculture in table-grape vineyards, we tested the overall grasping pipeline for detaching a (soft and fragile) table-grape berry from a bunch. 
As mentioned earlier, the challenge in manipulating these soft objects is to apply the necessary force to detach them, which typically varies based on the object, while avoiding any damage to the object. Examples of all manipulation tasks and of the integrated pipeline are provided in the supplementary video. 

For all experiments, we set the initial detection threshold equal to $\tau_{d,ini}=0.01$ and considered displacements of $0.001$~m to tighten the gripper in state 6. The noise thresholds $\tau_n^s$ were collected before each experiment using $K=100$ frames for each sensor, and the reference images $R^s$ were updated using $N=10$ frames. We set the closing gripper speed to $0.0015$~m/s. All experimental videos and tactile sensor images are available on the project website\footref{fn:website}.  

\subsection{Manipulation Algorithm Validation}

To validate the manipulation algorithm, we considered seven objects shown in Fig.~\ref{fig:objects}. In the figure, for each object (top row), an example of the respective tactile image is shown (bottom row). 
We performed two unplugging tasks of rigid connectors with different forces needed to unplug from their sockets and different
textures, one with a rough surface (Fig. \ref{fig:rough_connector}) and one with a smooth surface (Fig. \ref{fig:smooth_connector}). Next, we carried out 
lifting tasks on two fragile objects, a plastic glass (Fig. \ref{fig:glass}) and a raw egg (Fig. \ref{fig:egg}). 
Then, we realized detaching tasks of three soft fragile objects, that are tomatoes (Fig. \ref{fig:tomato}) and white  (Fig. \ref{fig:whitegrape}), and black (Fig. \ref{fig:blackgrape}) grapes. For these soft objects, we attached the respective connecting branch to the experimental setup in order to generate a realistic detaching force. 
In all tests, we started from a pre-grasping pose and required the robot to move up to perform the unplugging, lifting, or detaching task. Each manipulation task of an object was repeated three times. 
The results for all objects are provided in Table~\ref{tab:results}. Specifically, we report the average and standard deviation of  duration, compression percentage with respect to the gripper width at touch time, and the number of slip detections.

\begin{table}[]
    \centering
    \resizebox{1\linewidth}{!}{
    \begin{tabular}{|l|l|l|l|}
\hline Experiment & Duration [s]& $\%$ Compression & $\#$ Slippages \\ \hline
Rough conn. & $108.92 \pm 1.43$ & $45.45 \pm 1.27$ & $11.00 \pm 0.00$\\ \hline
Smooth conn. & $79.88 \pm 3.77$ & $10.47 \pm 1.28$ & $5.00 \pm 0.82$\\ \hline
Egg & $68.37 \pm 2.20$ & $3.69 \pm 0.32$ & $1.00 \pm 0.00$\\ \hline
Glass & $51.52 \pm 1.51$ & $1.67 \pm 0.74$ & $0.67 \pm 0.47$\\ \hline
Tomato & $43.50 \pm 2.09$ & $4.78 \pm 2.31$ & $0.67 \pm 0.47$\\ \hline
White grape & $55.45 \pm 1.15$ & $13.79 \pm 3.75$ & $1.33 \pm 0.47$\\ \hline
Black grape & $58.38 \pm 1.52$ & $18.45 \pm 2.28$ & $2.67 \pm 0.47$\\ \hline
    \end{tabular}
    }
    \caption{Results in terms of duration, $\%$ Compression, and $\#$~Slippages. Each manipulation is repeated three times, and average and standard deviation values are reported.  }
    \label{tab:results}
\end{table}

We can observe an obvious correlation between the number of slippages and the duration and compression needed to manipulate a given object: the higher the number of slippages, the longer the task duration and the higher the compression. 
The rough connector is by far the object that needs the most force to be unplugged, and therefore 
is the one resulting in  the highest number of slippages (equal to $11$) since a  strong grasp must be achieved to succeed in the manipulation. Similarly, the highest object compression compared to the touch configuration (equal to $45\%$ on average), and longest duration (equal to $109$~s on average) are recorded. A much easier unplugging task is obtained with the smooth connector, resulting in $5$ slippages on average. This shows that objects of the same category can yield very different results based on their specific features and properties.

The plastic glass and the tomatoes, on the other hand, led to the least amount of slippage detected (below $0.7$ on average). Some tests were successful without requiring any adjustments to the gripper closure. 
This result was expected as, for the glass manipulation,  the glass surface is not highly slippery, and  there are no significant opposing forces that need to be overcome during the lifting process; for the tomato detaching, the connection of the fruit to the branch is relatively weak and, again, does not need to counteract high forces. However, in both cases, the proposed algorithm successfully accomplished the tasks without deforming or damaging the objects. A slightly higher number of slippages (equal to $1$) was recorded with the raw egg lifting, due to the higher weight and a more slippery surface compared to the plastic glass.  

The grapes were the most challenging objects we tested, as they are significantly softer than the elastomer material of the DIGIT sensor. As a result, minimal deformation of the sensor material occurs during grasping, while the grape itself can be significantly deformed. Additionally, grapes are connected to the bunch by pedicels, which need to be pulled apart to detach a grape berry. This process requires exerting a certain amount of force, which  must be carefully controlled not to cause the grape to get crushed and burst open. On average, slippage was detected $2.67$ times for black grapes and $1.33$ times for white grapes, due to the different properties of the specific grape varieties. However, in all tests of both grape varieties, the system was able to successfully pick the grape.

\begin{figure}[t]
\begin{subfigure}[b]{0.45\textwidth}
  \centering
  \includegraphics[width=1\linewidth]
  {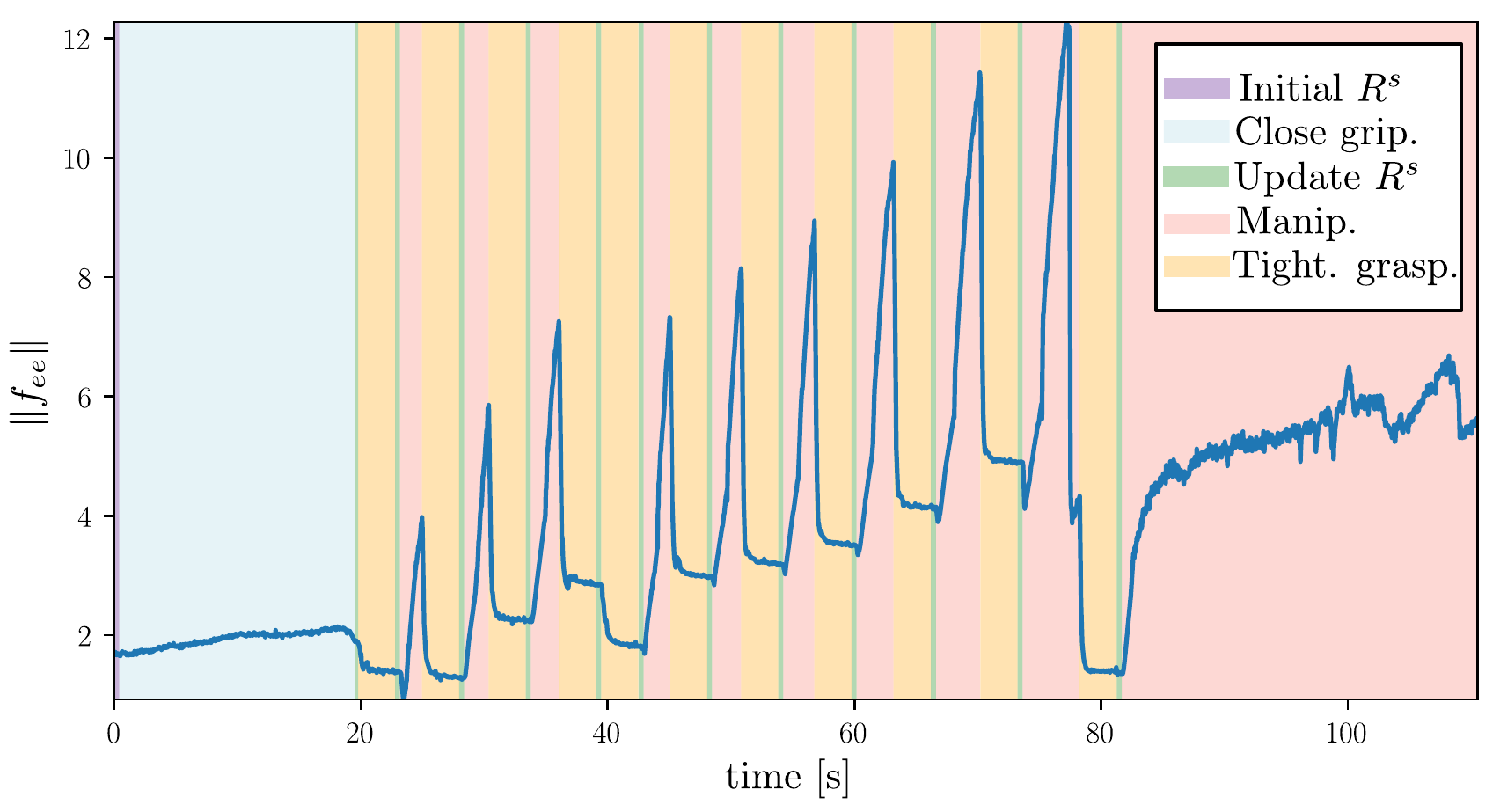}
    \caption{Rough connector. }
    \label{fig:force_connector_1}
\end{subfigure}
\begin{subfigure}[b]{0.45\textwidth}
  \centering
  \includegraphics[width=1\linewidth]
  {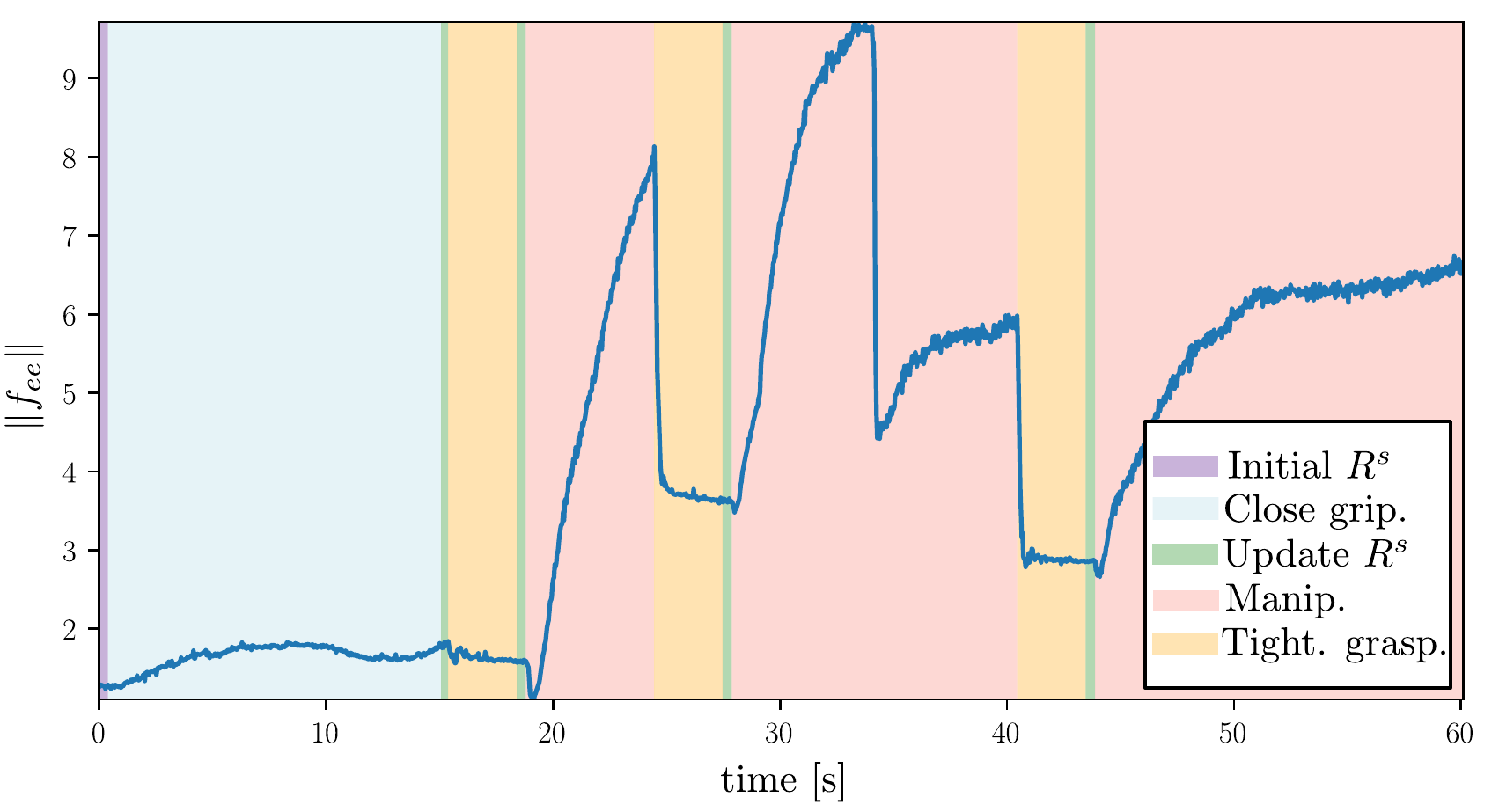}
    \caption{Black grape. }
    \label{fig:force_grape}
\end{subfigure}
\caption{Examples of the evolution of the force norm for manipulating the rough connector (top) and a black grape (bottom). The states  of the manipulation algorithm  are  also highlighted with different background colors.}
\label{fig:force}
\end{figure}

The norm of the estimated force at the end effector \cite{franka}, denoted by $\|f_{ee}\|$ and based on the joint torque sensor measures, is shown in  Fig.~\ref{fig:force}  for manipulating the rough connector (top) and a black grape (bottom). The states  of the manipulation algorithm are  also highlighted with different background colors. Some states are not visible due to short duration. 
The figure shows how the end effector force varies according to the specific phase of the manipulation algorithm. 
Specifically, for the  rough connector, every time a slip is detected (beginning of yellow background blocks), the grasp is tightened, leading to an increasing force generated to unplug the object during subsequent manipulation phases (light red background). This trend is observed until the connector is finally unplugged at about $t=80$~s. 
 Note that during the tightening phases, we stop pulling the object, which is why a drop in the force norm is observed. 
Regarding the black grape manipulation, we observe much fewer slippages compared to the previous case. However, the same trend is preserved: at the beginning of the first two manipulation phases, an increase in the force norm is observed to detach the grape until about $t=35$~s when the force drops since the grape is successfully detached.

\subsection{Integrated Pipeline Validation}
We now present the results obtained with the entire pipeline exploiting the suggestion of the grasping poses as described in Sec.~\ref{sec:integration}. The objective is to detach black grapes from a bunch, as shown in Fig. \ref{fig:real_bunch} and in the accompanying video. This application might be relevant, for instance, for assistive settings, and for quality inspection applications. In this case, during the manipulation phase, the robot was required to move horizontally in the opposite direction with respect to the bunch. For each grape detaching task, the  user cropped the point cloud to mainly contain the object and selected the reference grasp approach direction the sampled grasps should be close to. Given these inputs, the grasp sampler first generated grasp poses, and then a human manually selected the pose the robot should reach. 
We detached three consecutive grapes from a small bunch with an average duration of $111.93 \pm 12.14$ for the manipulation algorithm. The compression of the black grape in this setting was   $22.23 \pm 7.72$, which is slightly higher than the compression observed in the single grape setting. Similarly,  we also recorded a higher number of slippage occurrences ($4.67 \pm 3.77$),  which we attribute to the fact that the achieved grasps are not as stable as in the single grape setting. Nevertheless, our method was able to detach the grapes without damaging them in all cases\footref{fn:website}.

%% file: includes/discussion_conclusion.tex
\section{Discussion}

In this work, we introduced a simple-to-deploy algorithm detecting touch and slippage using the vision-based DIGIT sensors for a wide range of soft and rigid objects. This capability is essential for various manipulation tasks, as highlighted in our demonstration of unplugging, lifting, and detaching tasks.

The use of camera-based vision sensors, like the DIGIT, generates images that can be processed by advanced machine learning techniques, as discussed in related work and demonstrated in \cite{lambeta2020digit}, to enable touch and slip detection capabilities. Specifically, the Pytouch library \cite{Lambeta2021PyTouch} that is recommended for DIGIT use provides pre-trained machine learning models which can be utilized for touch detection. 
However, in our test, we could not reliably detect the object as shown at the link\footnote{\url{https://vision-tactile-manip.github.io/exp/#pytouch}}. We hypothesize that this is related to the fact that pre-trained models might be hardware specific. This motivates why we opted for a learning-free approach to touch and slip detection which can be easily adapted to different settings. 

Another challenge that we encountered is that the quality of the tactile images, in terms of sharpness of the manipulated object, increases as the contact force increases, i.e., the harder we press, the more details are visible. 
This is not an issue for rigid objects, as impressively demonstrated in \cite{lambeta2020digit} (and shown in this work on the rough AUX connector), where we can apply relatively high forces without damaging the object. 
However, it represents a major limitation when dealing with soft objects. In  cases where the object is significantly softer than the sensor elastomer material, we might not be able to observe any deformation through the images,  even if the object is being compressed. 
This motivates why we designed an increasing detection threshold which allows us to detect small changes at first (which are suitable for soft objects), but if needed it also allows us to apply significant forces after few iterations (which are suitable for rigid objects). 

However, as a drawback of the proposed approach, we identify the fact that it may result in a rather slow manipulation process, since we need to iteratively detect slip occurrences in order to modulate the gripper closure. In contrast, humans can easily identify in advance properties of the objects, which enable their manipulation without the need of multiple slip occurrences. Therefore, we aim to improve this aspect in future work by inferring relevant object properties while performing manipulation and exploiting them to achieve a more proactive behavior.

\section{Conclusion}

In this work, we proposed a manipulation algorithm for soft and rigid objects using vision-based tactile sensors mounted on the robot gripper. The algorithm 
is based on a touch and slip detection method which evaluates the variations in the sensor images and relies on a single hyperparameter. No knowledge about  the shape, texture, and physical properties of the object to manipulate was required. 
We showed the effectiveness of the approach on seven real-world objects, including rigid, soft, and fragile ones, using two DIGIT sensors for unplugging, lifting, and detaching tasks. Finally, we combined the manipulation strategy with a grasping proposal system to provide a comprehensive framework and validated it on a grape detaching task. 